\documentclass{article}
\usepackage{spconf,amsmath,epsfig}
\usepackage{amsfonts}
\renewcommand{\paragraph}[1]{\noindent\textbf{#1}\quad}
\newtheorem{theorem}{Theorem}
\DeclareMathOperator{\trace}{tr}
\usepackage{booktabs}
\usepackage{xcolor}
\usepackage{multirow}
\usepackage{color, colortbl}
\definecolor{LightCyan}{rgb}{0.99,0.84,0.69}
\usepackage{cleveref}
\usepackage{soul}
\usepackage[font=small]{caption}
\usepackage{subcaption}
\usepackage{floatrow}
\newfloatcommand{capbtabbox}{table}[][\FBwidth]
\usepackage[linesnumbered,ruled,vlined]{algorithm2e} %

\let\OLDthebibliography\thebibliography
\renewcommand\thebibliography[1]{
  \OLDthebibliography{#1}
  \setlength{\parskip}{0pt}
  \setlength{\itemsep}{0pt plus 0.3ex}
}

\begin{document}\sloppy

\def\x{{\mathbf x}}
\def\L{{\cal L}}

\title{M\lowercase{ulti}-D\lowercase{imensional} M\lowercase{odel} C\lowercase{ompression} \lowercase{of} V\lowercase{ision} T\lowercase{ransformer}}
%
\name{Zejiang Hou, Sun-Yuan Kung \vspace{-0.1in}}
\address{Princeton University \\ {\tt\small \{zejiangh,kung\}@princeton.edu}}

\maketitle

\begin{abstract}
Vision transformers (ViT) have recently attracted considerable attentions, but the huge computational cost remains an issue for practical deployment.
Previous ViT pruning methods tend to prune the model along one dimension solely, which may suffer from excessive reduction and lead to sub-optimal model quality.
In contrast, we advocate a multi-dimensional ViT compression paradigm, and propose to harness the redundancy reduction from attention head, neuron and sequence dimensions jointly.
We firstly propose a statistical dependence based pruning criterion that is generalizable to different dimensions for identifying deleterious components.
Moreover, we cast the multi-dimensional compression as an optimization, learning the optimal pruning policy across the three dimensions that maximizes the compressed model's accuracy under a computational budget. The problem is solved by our adapted Gaussian process search with expected improvement.
Experimental results show that our method effectively reduces the computational cost of various ViT models.
For example, our method reduces 40\% FLOPs without top-1 accuracy loss for DeiT and T2T-ViT models, outperforming previous state-of-the-arts.
\end{abstract}
\section{Introduction}
\label{sec:intro}
Vision transformers (ViT) \cite{dosovitskiy2020image,touvron2021training} have achieved substantial progress in prevalent computer vision tasks such as image classification, object detection and semantic segmentation.
However, ViT models suffer from excessive computational and memory cost, impeding their deployment in resource-restricted, real-time, or low-powered applications.
Although a plethora of model compression algorithms 
have been proposed for convolutional neural networks (CNN), it is not immediately clear whether they are the same effective for vision transformers and there are only few works \cite{pan2021ia,chen2021chasing,xu2021evo,rao2021dynamicvit,tang2021patch} on accelerating vision transformers.

Prior works \cite{child2019generating,wang2020linformer,xiong2021nystr} in natural language processing (NLP) focus on addressing the quadratic complexity of the softmax-attention in transformer (presumably because the input has very long sequence in NLP tasks). However, in ViT models softmax-attention constitutes small fraction of the total FLOPs as shown in the table below.
Instead, the projection layers are the major computation bottleneck, and the complexity of these layer are affected by the attention head, neuron, and sequence dimensions.

\begingroup
\setlength{\tabcolsep}{1.7pt}
\begin{table}[h]
\vspace{-0.1in}
\centering
\footnotesize
\begin{tabular}{l | c | c | c | c}
    Model & Softmax-attention & MHSA projections & FFN projections & Total\\\hline
    DeiT-S \cite{touvron2021training} & 0.36G (8\%) & 1.39G (30\%) & 2.79G (61\%) & 4.6G \\
    DeiT-B \cite{touvron2021training} & 0.72G (4\%) & 5.58G (32\%) & 11.15G (64\%) & 17.5G \\
\end{tabular}
\vspace{-0.1in}
\label{tab:vit_computation_analysis}
\end{table}
\endgroup

ViT models split the input image into a sequence of image tokens.
Considering that not all tokens contribute to the final predictions \cite{rao2021dynamicvit} and the high similarity between tokens within a layer \cite{tang2021patch}, 
recent ViT compression methods \cite{rao2021dynamicvit,xu2021evo,tang2021patch} focus on unstructured token pruning by removing the redundant and unimportant tokens.
Since the self-attention operator can process variable sequence length, 
this unstructured sequence reduction brings about practical acceleration.
However, to achieve more compelling computational cost reduction, excessive pruning of this single sequence dimension leads to unacceptable accuracy loss as discussed in \Cref{sec:ablation}.
This motivates our study on how to find an optimal compression policy to reduce the computational cost from multiple dimensions (i.e, head, neuron, and sequence) jointly, thus achieving better computation-accuracy trade-off.

Multi-dimensional compression is challenging.
Most pruning algorithms are designed by the unique property of a single dimension, e.g., using column mean of the attention matrix for sequence reduction. 
They may hardly generalize to other dimensions.
Moreover, the large decision space stemming from three integrated dimensions makes it hard to decide how much of each dimension should be compressed.

\paragraph{Our contributions.} To rectify the aforementioned problems, we firstly propose a general data-aware pruning criterion that is applicable to both structured neuron or head reduction and unstructured sequence reduction. 
The criterion measures the statistical dependency between the features of a dimension and the output predictions of the model based on the Hilbert-Schimdt norm of the cross-variance operator.
Moreover, we formulate the multi-dimensional compression as an optimization problem, seeking the optimal pruning policy (i.e, pruning ratios across the three dimensions) that maximize the compressed model's accuracy under a target computational cost.
Considering the non-differentiability of the problem 
and the optimization efficiency,
we propose to use Gaussian process (GP) search with expected improvement to estimate the compressed model's accuracy for different pruning policies,
thus transform the problem to a simpler and solvable non-linear programming.
To fit the GP model, we need to evaluate the actual accuracy corresponding to a small set of sampled pruning policies.
We further design a weight sharing mechanism 
for fast accuracy evaluation without training each compressed model from scratch.
Our method exhibits superior accuracy compared to previous state-of-the-art ViT pruning methods under same FLOPs reductions.
When compressing DeiT \cite{touvron2021training} and T2T-ViT \cite{Yuan_2021_ICCV} on ImageNet \cite{imagenet_cvpr09}, 
our method reduces 40\% $\sim$ 60\% FLOPs and yields 1.3$\times$ $\sim$ 2.2$\times$ practical speedup without significant accuracy drop.

\vspace{-0.1in}
\section{Related Works}
\label{sec:related_works}
\vspace{-0.1in}

\paragraph{Vision transformer model compression.}
To improve the efficiency of ViT models,
\cite{zhu2021visual, chen2021chasing} applies structured neuron pruning or unstructured weight pruning.
\cite{rao2021dynamicvit,xu2021evo,tang2021patch,pan2021ia} applies dynamic or static token sparsification.
\cite{touvron2021training} proposes a knowledge distillation method specific to transformer by introducing a distillation token.
\cite{liu2021post} uses post-training quantization to reduce the model size. 
However, multi-dimensional compression of ViT models has been rarely explored, and its effectiveness compared to uni-dimensional compression is unknown.
In this work, we will show that excavating the redundancy from multiple dimensions is imperative to achieve more appealing FLOPs reductions, and our method achieves state-of-the-art pruning results compared to previous methods. 


\paragraph{Multi-dimensional pruning} methods have been proposed for compressing CNNs.
\cite{wen2016learning,huang2018data,lin2019towards} impose sparsity regularization, e.g., group LASSO, to prune channels and layers in CNNs. \cite{guo2020multi} uses L1 regularization to prune channels and feature-map spatial sizes in CNNs.
In addition to the regularization-based methods, \cite{liu2020joint, wang2021accelerate} directly search the number of channels, layers and spatial sizes under a FLOPs budget by reinforcement learning or polynomial regression.
In contrast, our method is specially designed for ViT compression by jointly pruning heads in MHSA, neurons in FFN, and sequence. We propose a dependency based pruning criterion and an efficient Gaussian process search to learn the optimal compression policy.

\vspace{-0.1in}
\section{Methodology}
\label{sec:methodology}
\vspace{-0.1in}

\subsection{Preliminary}
\vspace{-0.1in}
ViT model contains interleaved multi-head self-attention (MHSA) and feed-forward network (FFN) modules. Denote the input features to MHSA and FFN in the $l^{\text{th}}$ layer by $X^l,Z^l\in\mathbb{R}^{N \times d}$, where $N$ is the sequence length and $d$ is the embedding dimension. The MHSA module has operations:\\
    \centerline{$\text{MHSA}(X^l) = \sum_{h=1}^H\text{softmax}(\dfrac{Q_hK_h^T}{\sqrt{d_h}})V_hW^o$} \\
where the query, key and value are computed by $Q_h=X^lW^Q_h, K_h=X^lW^K_h, V_h=X^lW^V_h$ ($W^Q_h,W^K_h,W^V_h\in\mathbb{R}^{d\times d_h}$), $H$ is the number of heads, and $W^o\in\mathbb{R}^{d_h\times d}$ is output projection. The FFN module has operations:\\
    \centerline{$\text{FFN}(Z^l) = \sigma(Z^lW_1+b_1)W_2+b_2$} \\
where $\sigma$ denotes the activation function and $W_1,W_2^T\in\mathbb{R}^{d\times d'}$, $b_1\in\mathbb{R}^{d'},b_2\in\mathbb{R}^{d}$ are projection matrices and biases.

\vspace{-0.1in}
\subsection{Dependency based pruning criterion}
\label{sec: dependency pruning}
\vspace{-0.1in}

Our goal is to accelerate ViT models by pruning multiple dimensions jointly, including number of neurons in FFN, heads in MHSA, and the sequence length.
We need a general criterion that can identify the deleterious features in different dimensions. 
Intuitively, the unimportant features contributes least to the output predictions. 
In other words, the output of the model has weak dependency on the unimportant features. Thus, we propose dependency based pruning, which evaluates the importance based on statistical dependency between the features and the output predictions of the model.

Denote the random variable of the features by $Z$ and the random variable of the model outputs by $Y$. 
Let $P_{Z,Y}$ be the joint distribution.
To measure the dependence between $Z$ and $Y$, we define the cross-covariance operator based on \cite{baker1973joint}:
\begin{equation}
C_{zy} := \mathbb{E}_{zy}[(\Phi(z)-\mu_{z})\otimes(\Psi(y)-\mu_{y})]
\end{equation}
where $\Phi~(\text{or}~\Psi)$ defines a kernel mapping from the feature space (or model output space) to a reproducing kernel Hilbert space (RKHS), with mean vector $\mu_z$ (or $\mu_y$).
$\otimes$ denotes the tensor product. 
To summarize the degree of dependence between $Z$ and $Y$, we use Hilbert-Schmidt norm of the cross-covariance operator $C_{zy}$, which is given by the trace of $C_{zy}C_{zy}^T$, 
because it can detect non-linear dependency as shown in the following theorem.
\begin{theorem}[\cite{gretton2005measuring}]
\textit{Given RKHSs with characteristic kernels. Then, $\|C_{zy}\|_{HS}^2=0$ if and only if Z and Y are independent.}
\end{theorem}
Characteristic kernels, e.g., Gaussian kernel $k(x,x')=\text{exp}\big(-\|x-x'\|_2^2/(2\sigma^2)\big)$, 
allows us to measure arbitrary mode of dependence between $Z$ and $Y$.
Features with high $\|C_{zy}\|_{HS}^2$ have high dependency with the outputs of the model, indicating that the features may have considerable influence to the output predictions, thus they should be retained.

To use the dependency criterion in practice, we need an empirical estimate from a batch of training samples.
Denote the kernel functions by $k(z,z')$ and $l(y,y')$. Let $\mathbf{K},\mathbf{L}$ be the Gram matrices ($\mathbf{K}_{i,i'}=k(z_i,z_{i'})$, $\mathbf{L}_{i,i'}=l(y_i,y_{i'})$) defined over the features and model outputs of $B$ training samples.
An empirical estimator of $\|C_{zy}\|_{HS}^2$ is given by:
\begin{equation}
\widehat{\|C_{zy}\|_{HS}^2} := (B-1)^{-2}\trace({\mathbf{K}\mathbf{C}\mathbf{L}\mathbf{C}})\label{empirical_dependency}
\end{equation}
where $\mathbf{C}=\mathbf{I}_B-({1}/{B})\mathbf{1}_B\mathbf{1}_B^T$ is the centering matrix 
($\mathbf{I}_B,\mathbf{1}_B$ represent identity matrix and all-ones vector) 
and $\trace(\cdot)$ is the matrix trace operation. 
As shown in \cite{gretton2005measuring}, this empirical estimator converges sufficiently: with high probability, $|\widehat{\|C_{zy}\|_{HS}^2} - \|C_{zy}\|_{HS}^2|$ is bounded by a very small constant.

\paragraph{FFN neuron reduction.}
We prune neurons in the intermediate layer of FFN modules. 
Denote the features from the intermediate layer in the $l^{\text{th}}$ FFN by $Z^l\in\mathbb{R}^{B\times N \times d'}$, where each neuron $j\in[d']$ has features $Z^l_{:,:,j}\in\mathbb{R}^{B\times N}$.
Given a neuron pruning ratio $\kappa^l$, we retain $\lceil (1-\kappa^l)d'\rceil$ important neurons.
We compute the dependency score for each neuron in the layer by $\psi_j^l=\trace({\mathbf{K}^l_j\mathbf{C}\mathbf{L}\mathbf{C}})$, where $\mathbf{K}_j^l\in\mathbb{R}^{B\times B}$ is the Gram matrix defined over $Z^l_{:,:,j}$, i.e., $[\mathbf{K}_j^l]_{i,i'} = k(Z^l_{i,:,j},Z^l_{i',:,j})$. Then, we rank the dependency scores in descending order, and identify the important neurons by $\text{ArgTopK}(\{\psi_1^l...\psi_{d'}^l\};\lceil (1-\kappa^l)d'\rceil)$, which gives the neuron indices with the top $\lceil (1-\kappa^l)d'\rceil$ dependency scores. 
The remaining bottom-ranking neurons are pruned.

\paragraph{Attention head reduction.}
Denote the output features of the self-attention operator in the $l^{\text{th}}$ MHSA by $Z^l\in\mathbb{R}^{B\times N \times H \times d_h}$. 
Given a head pruning ratio $\zeta^l$, the head pruning procedure is similar to the neuron pruning procedure, except that each head $h\in[H]$ has output features $Z^l_{:,:,h,:}\in\mathbb{R}^{B\times N \times d_h}$. 
To construct the feature Gram matrix, we 
perform mean pooling along the embedding dimension to obtain $\tilde{Z}^l_{:,:,h}=\text{mean}(Z^l_{:,:,h,:};\text{dim=-1})$ ($\text{-1}$ means the last tensor dimension), 
which has a size of $\mathbb{R}^{B\times N}$. 
Then, we compute the Gram matrix $\mathbf{K}^l_h$ over $\tilde{Z}^l_{:,:,h}$ and the head dependency score $\psi^l_h$. We retain heads with the top $\lceil (1-\zeta^l)H\rceil$ dependency scores and prune the bottom-ranking heads.

\paragraph{Sequence reduction.}
To achieve unstructured sequence reduction, we insert token selection layer (TSL) (no extra parameters) after the MHSA module and before the FFN module at each transformer layer. 
Let $Z^l\in\mathbb{R}^{B\times N\times d}$ be the input features to $l^\text{th}$ TSL. 
Given a sequence reduction ratio $\nu^l$, TSL outputs the selected $\lceil (1-\nu^l)N \rceil$ important tokens from $Z^l\in\mathbb{R}^{B\times N\times d}$ according to indices $\text{ArgTopK}(\{\psi_1^l...\psi_{N}^l\};\lceil (1-\nu^l)N\rceil)$.
That is, TSL extracts a sub-tensor from the input features, the output has size $\mathbb{R}^{B\times \lceil (1-\nu^l)N \rceil \times d}$. 
The token dependency scores $\psi_n^l,n\in[N]$ are computed based on the Gram matrix $\mathbf{K}^l_n$ defined over the token features $Z^l_{:,n,:}\in\mathbb{R}^{B\times d}$.
After TSL, the sequence length in subsequent layers becomes $\lceil (1-\nu^l)N \rceil$. And the subsequent TSLs select tokens from what are preserved by the previous TSLs.

\vspace{-0.1in}
\subsection{Multi-dimensional pruning via Gaussian Process}
\vspace{-0.1in}

\paragraph{Formulation.}
We aim to strike a balance among the three pruning dimensions so that the compressed model has the best accuracy under a computation constraint. Thus, our multi-dimensional compression is formulated as:
\begin{align}
& \underset{\{\kappa^l,\zeta^l,\nu^l\}_{l=1}^L}{\text{max}}\text{Accuracy}(\{\kappa^l,\zeta^l,\nu^l\}_{l=1}^L) \label{mdp_formulation} \\
& \hspace{0.2in} \text{s.t.}~~\mathcal{C}(\{\kappa^l,\zeta^l,\nu^l\}_{l=1}^L) \leq \mathcal{T}\nonumber
\end{align}
where $\text{Accuracy}(\cdot)$ gives the accuracy of the compressed model with pruning ratios $\{\kappa^l,\zeta^l,\nu^l\}_{l=1}^L$, 
$\mathcal{C}(\cdot)$ represents the FLOPs of the compressed model, and $\mathcal{T}$ is the constraint.

\paragraph{Gaussian process search.}
To solve problem \eqref{mdp_formulation}, we resort to Bayesian optimization, as it provides an efficient framework for objectives that may not be differentiable or expressed in a closed-form.
We use Gaussian process (GP) \cite{rasmussen2003gaussian} with expected improvement (EI) to estimate the accuracy function in closed form, so that problem \eqref{mdp_formulation} is transformed to a simpler and solvable constrained non-linear optimization.

For simplicity, we denote a pruning policy by $\omega=\{\kappa^l,\zeta^l,\nu^l\}_{l=1}^L$.
A Gaussian process is described by: \\
\centerline{$f \sim \mathcal{G}\mathcal{P}(\mu(\cdot), k(\cdot,\cdot))$} \\
where the GP has a mean function $\mu(\omega) = \mathbb{E}[f(\omega)]$ and a covariance kernel $k(\omega,\omega') = \mathbb{E}[(f(\omega)-\mu(\omega))(f(\omega')-\mu(\omega'))]$.
We sample $m$ different pruning policies $\Omega=\{\omega_i\}_{i=1}^m$ satisfying the constraint $\mathcal{T}$, obtain $m$ compressed models using our dependency based pruning, evaluate their actual accuracy $\mathcal{A}(\omega_i)$ on a hold-out set, and fit the GP model by points $\{\omega_i,\mathcal{A}(\omega_i)\}_{i=1}^m$. 
At a new pruning policy $\hat{\omega}$, the posterior of
$f$ at this point is given by $\hat{f}\sim\mathcal{N}(\hat{\mu},\hat{\Sigma}),\hat{\mu}=\mu(\hat{\omega})+k(\hat{\omega},\Omega)k(\Omega,\Omega)^{-1}(\mathcal{A}(\Omega)-\mu(\Omega)),\hat{\Sigma}=k(\hat{\omega},\hat{\omega})-k(\hat{\omega},\Omega)k(\Omega,\Omega)^{-1}k(\Omega,\hat{\omega})$.
The expected improvement at $\hat{\omega}$ is computed in closed form by:\\
\centerline{\resizebox{0.9\hsize}{!}{$
    \text{EI}(\hat{\omega}) = (\mathcal{A}^*-\hat{\mu})\Phi\big( {(\mathcal{A}^*-\hat{\mu})}/{\hat{\Sigma}} \big)+\hat{\Sigma}\phi\big( {(\mathcal{A}^*-\hat{\mu})}/{\hat{\Sigma}} \big)
$}} \\
where $\Phi,\phi$ represent the CDF and PDF of the standard normal distribution, and $\mathcal{A}^*$ is the accuracy of the best policy in $\Omega$.
Therefore, the most promising policy $\omega^*$ to evaluate is the solution to the following non-linear programming:
\begin{equation}
    \underset{\hat{\omega}}{\text{max}}~~\text{EI}(\hat{\omega}),~~\text{s.t.}~~\mathcal{C}(\hat{\omega})\leq\mathcal{T}\label{EI_maximization}
\end{equation}
Since both the objective and constraints \footnote{FLOPs of a ViT model can be computed by closed-form formula with the pruning ratios in three dimensions.} in \eqref{EI_maximization} have closed-form formulas,
the problem can be solved by standard constrained optimization solver, and we use
sequential quadratic programming (SQP) \cite{kraft1988software}.
We iterate the search process by using the obtained $\omega^*$ and its actual accuracy $\mathcal{A}(\omega^*)$ to refine the GP model, and find the next (more optimal) policy until no accuracy improvement is observed.
With the final optimal pruning policy, we apply our dependency based pruning, and finetune the compressed model.

\paragraph{Accuracy evaluation in GP search.}
Our GP search involves evaluating the actual accuracy of compressed models.
Instead of training many models with different pruning policies from scratch, we apply weight sharing \cite{guo2020single} for efficient accuracy evaluation. 
We pre-train the full model weights $\mathcal{W}$ in a way that the accuracy of sub-models with inherited weights are predictive for the accuracy obtained by training them independently. Our pre-training objective is given by:
\begin{equation}
\underset{\mathcal{W}}{\text{min}}~~\mathbb{E}_{(x,y)\sim\mathcal{D},\omega\sim\textit{U}_{\mathcal{T}}}[\mathcal{L}(y|x;\mathcal{W}(\omega))]\label{weight_sharing}
\end{equation}
where $\mathcal{D}$ is the training set, $\textit{U}_\mathcal{T}$ is a constrained uniform sampling distribution \footnote{We sample the pruning policy repeatedly until the compressed model FLOPs satisfies the constraint $\mathcal{T}$.} of pruning policies, and $\mathcal{L}$ is the loss function.
$\mathcal{W}(\omega)$ means selecting weights from $\mathcal{W}$ to form the compressed model with pruning policy $\omega$.
For neuron and head dimensions, we keep the weights corresponding to neurons or heads with top dependency scores as described in \Cref{sec: dependency pruning}.
Reducing sequence dimension
does not require modification to the model weights, 
thus all parameters are shared.
Due to weight sharing, one only need to train one set of weights and directly evaluate the accuracy of different pruning policies by inheriting weights from the full model.
This takes much less time than training each compressed model from scratch and makes our GP search highly efficient.


\begingroup
\setlength{\tabcolsep}{4pt}
\begin{table}[t]
\centering
\footnotesize
\begin{tabular}{l c c c c}\toprule
    \textbf{Method} & \textbf{Top-1} & \textbf{Top-1 drop} & \textbf{FLOPs} & \textbf{FLOPs reduction} \\\midrule
    \multicolumn{5}{l}{\textit{\textbf{DeiT-Small model}}} \\
    \;\;Baseline \cite{touvron2021training} & 79.8\% & - & 4.6G & - \\
    \;\;SPViT \cite{he2021pruning} & 78.3\% & 1.5\% & 3.3G & 29\% \\
    \;\;IA-RED$^2$ \cite{pan2021ia} & 79.1\% & 0.7\% & 3.1G & 32\% \\
    \;\;S$^2$ViTE \cite{chen2021chasing} & 79.2\% & 0.6\% & 3.1G & 32\% \\
    \;\;Evo-ViT \cite{xu2021evo} & 79.4\% & 0.4\% & 2.9G & 37\% \\
    \;\;DynamicViT \cite{rao2021dynamicvit} & 79.3\% & 0.5\% & 2.9G & 37\% \\
    \rowcolor{LightCyan} \;\;Ours & 79.9\% & -0.1\% & 2.9G & 37\% \\
    \;\;UVC \cite{anonymous2021unified} & 78.4\% & 1.4\% & 2.4G & 48\% \\
    \rowcolor{LightCyan} \;\;Ours & 79.3\% & 0.5\% & 1.8G & 60\% \\
    \midrule
    \multicolumn{5}{l}{\textit{\textbf{DeiT-Base model}}} \\
    \;\;Baseline \cite{touvron2021training} & 81.8\% & - & 17.5G & - \\
    \;\;VTP \cite{zhu2021visual} & 81.3\% & 0.5\% & 13.8G & 22\% \\
    \;\;IA-RED$^2$ \cite{pan2021ia} & 80.3\% & 1.5\% & 11.8G & 33\% \\
    \;\;S$^2$ViTE \cite{chen2021chasing} & 82.2\% & -0.4\% & 11.8G & 33\%  \\
    \;\;SPViT \cite{he2021pruning} & 81.6\% & 0.2\% & 11.7G & 33\% \\
    \;\;Evo-ViT \cite{xu2021evo} & 81.3\% & 0.5\% & 11.7G & 33\% \\
    \;\;DynamicViT \cite{rao2021dynamicvit} & 81.3\% & 0.5\% & 11.2G & 36\% \\
    \rowcolor{LightCyan} \;\;Ours & 82.3\% & -0.5\% & 11.2G & 36\% \\
    \;\;UVC \cite{anonymous2021unified} & 80.6\% & 1.2\% & 8.0G & 55\% \\
    \rowcolor{LightCyan} \;\;Ours & 81.5\% & 0.3\% & 7.0G & 60\% \\
    \midrule
    \multicolumn{5}{l}{\textit{\textbf{T2T-ViT-14 model}}} \\
    \;\;Baseline \cite{Yuan_2021_ICCV} & 81.5\% & - & 4.8G & - \\
    \;\;PatchSlim \cite{tang2021patch} & 81.1\% & 0.4\% & 2.9G & 40\% \\
    \rowcolor{LightCyan} \;\;Ours & 81.7\% & -0.2\% & 2.9G & 40\% \\
    \bottomrule
\end{tabular}
\caption{
Comparison of our compressed ViT models versus baselines and previous methods on ImageNet.
Negative ``Top-1 drop'' means that our accuracy improves over the baseline.
\vspace{-0.1in}
}
\label{tab:main_results}
\vspace{-0.1in}
\end{table}
\endgroup

\vspace{-0.1in}
\section{Experiments}
\label{sec:experiments}
\vspace{-0.1in}

\subsection{Setup}
\vspace{-0.1in}

We conduct experiments on the ImageNet dataset \cite{imagenet_cvpr09} with represntative ViT models, DeiT \cite{touvron2021training} and T2T-ViT \cite{Yuan_2021_ICCV}, which were also used by previous ViT compression methods \cite{rao2021dynamicvit,chen2021chasing,pan2021ia,xu2021evo,tang2021patch}.
All experiments run on PyTorch framework with Nvidia A100 GPUs.
We firstly pre-train the models from scratch with Eq.\eqref{weight_sharing}, and follow the same training hyper-parameters as the paper of DeiT and T2T-ViT.
Then, we conduct Gaussian process (GP) search
to obtain the optimal pruning policy.
The target computational costs are listed as the FLOPs reductions in \Cref{tab:main_results}.
The initial population size to fit a GP model is 100, and the GP search runs for 100 iterations.
We randomly sample 50k images (50 images per class) from the training set of ImageNet for accuracy evaluation during GP search. 
Our GP search process is computationally efficient, taking less than 1 hour on a single A100 GPU for all cases.
Based on the optimal pruning policy, we compress the pre-trained models along head, neuron and sequence dimensions using our dependency based pruning.
The compressed model is finetuned following the same training strategy as \cite{touvron2021training,tang2021patch}.
To compute the dependency score in Eq.\eqref{empirical_dependency}, 
we randomly sample a mini-batch of 256 images from the training set and use Gaussian kernel with bandwidth $\sigma=1$.
In contrast to \cite{rao2021dynamicvit,anonymous2021unified},
our method does not use any knowledge distillation technique.
More implementations and results are put in the Appendix.

\begingroup
\setlength{\tabcolsep}{8.7pt}
\begin{table}[t]
\centering
\footnotesize
\begin{tabular}{l | c c | c c}\toprule
    \multirow{2}{*}{\textbf{Method}} & \multicolumn{2}{c}{\textbf{DeiT-Ti}} & \multicolumn{2}{c}{\textbf{DeiT-S}} \\
    & \textbf{Top-1} & \textbf{FLOPs} & \textbf{Top-1} & \textbf{FLOPs} \\\midrule
    \;Baseline & 72.2\% & 1.3G & 79.8\% & 4.6G \\
    \;Baseline$_{2\times}$ & 73.9\% & 1.3G & 81.0\% & 4.6G \\
    \;Manifold$^\dagger$ \cite{jia2021efficient} & 75.1\% & 1.3G & 81.5\% & 4.6G \\
    \;UP-DeiT$^\dagger$ \cite{yu2021unified} & 75.8\% & 1.3G & 81.6\% & 4.6G \\
    \;NViT-DeiT$^\dagger$ \cite{yang2021nvit} & 76.2\% & 1.3G & 82.1\% & 4.6G \\
    \;\textbf{Ours} & \textbf{77.0\%} & 1.3G & \textbf{82.1\%} & 4.6G \\
    \bottomrule
\end{tabular}
\caption{
Results of expand-then-compress on ImageNet.
``$2\times$'' means doubling the training epochs when training the original model.
$^\dagger$: methods using knowledge distillation.
\vspace{-0.1in}
}
\label{tab:expand_then_prune}
\vspace{-0.1in}
\end{table}
\endgroup

\vspace{-0.1in}
\subsection{Comparison with the state-of-the-art methods.}
\vspace{-0.1in}

We compare with the latest ViT model compression methods, including sequence reduction methods (DynamicViT \cite{rao2021dynamicvit}, IA-RED$^2$ \cite{pan2021ia}, PatchSlim \cite{tang2021patch}, Evo-ViT \cite{xu2021evo}), weight pruning methods (VTP \cite{zhu2021visual}, S$^2$ViTE \cite{chen2021chasing}), unified ViT compression method UVC \cite{anonymous2021unified}, and NAS-based ViT pruning method \cite{he2021pruning}. The results are shown in \Cref{tab:main_results}.

Our method achieves noticeably higher accuracy than previous methods under same FLOPs. For example, our pruned DeiT-S model with 37\% FLOPs reduction outperforms DynamicViT and Evo-ViT by 0.6\% and 0.5\% accuracy, respectively. On the other hand, at the same target accuracy, our method achieves higher FLOPs reduction. For example, our pruned DeiT-B model yields 60\% FLOPs reduction with 81.5\% top-1 accuracy, compared to DynamicViT and Evo-ViT yielding less than 36\% FLOPs reduction. These results clearly evidence the advantage of pruning multiple dimensions in the ViT models, when we aim to achieve compelling FLOPs reduction without compromising too much accuracy.

\vspace{-0.1in}
\subsection{Expand-then-compress improves existing model.}
\vspace{-0.1in}

Apart from compressing the model for faster inference, our method can improve existing 
models for higher accuracy under the same FLOPs (\Cref{tab:expand_then_prune}).
More exactly, we apply our method to compress a scaled-up DeiT-Ti model (width scaled by $2\times$), with the goal of reducing its FLOPs to the same level as the original DeiT-Ti. 
Notably, the obtained model achieves 77\% top-1 accuracy at 1.3 GFLOPs, outperforming the original DeiT-Ti (trained with longer training epochs) by 3.1\%.
Same phenomenon also applies to DeiT-S where our method achieves 82.1\% top-1 accuracy, improving the baseline by 1.1\%.
These results suggest that heavily compressed larger ViT models may achieve higher accuracy than small models.
In summary, our method achieves better Pareto frontier compared to existing models as shown in Figure 2.


\begingroup
\setlength{\tabcolsep}{1.5pt}
\begin{table*}[t]
\centering
\footnotesize
\begin{tabular}{l |c c c | c c}\toprule
    \textbf{Model} & \textbf{Neuron} & \textbf{Head} & \textbf{Sequence} & \textbf{Top-1} & \textbf{FLOPs} \\\midrule
    \multirow{7}{*}{DeiT-B}
    & - & - & - & 81.8\% & 17.5G \\
    & \checkmark & - & - & 79.4\% & 7.4G \\
    & - & - & \checkmark & 79.5\% & 7.0G \\
    & \checkmark & \checkmark & - & 80.4\% & 7.1G \\
    & \checkmark & - & \checkmark & 80.7\% & 7.0G \\
    & - & \checkmark & \checkmark & 80.2\% & 7.0G \\
    & \checkmark & \checkmark & \checkmark & \textbf{81.5\%} & 7.0G \\
    \bottomrule
\end{tabular}
\begin{tabular}{c}
\hspace{-0.09in}
\includegraphics[scale=0.22]{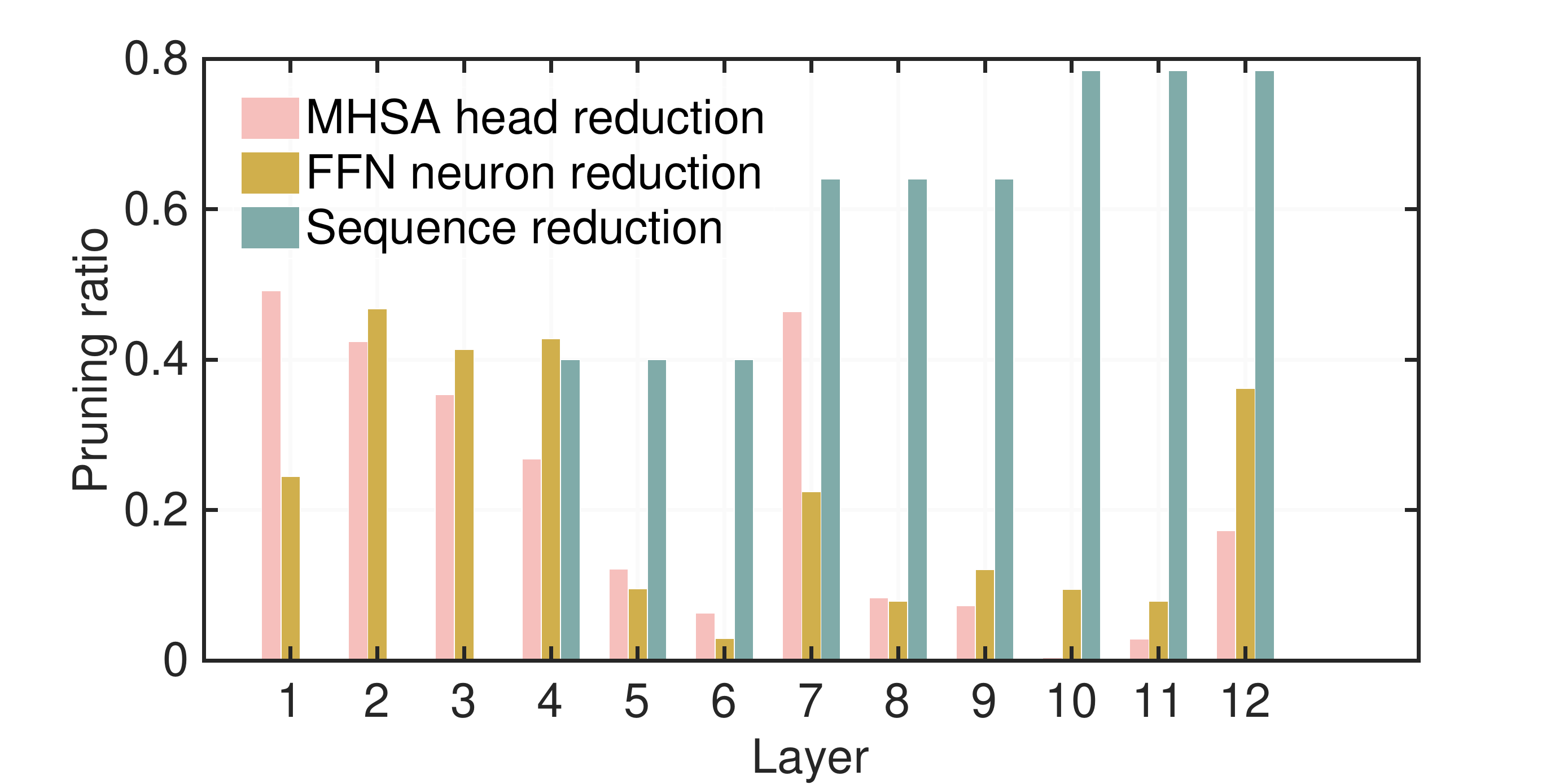}
\hspace{-0.25in}
\includegraphics[scale=0.22]{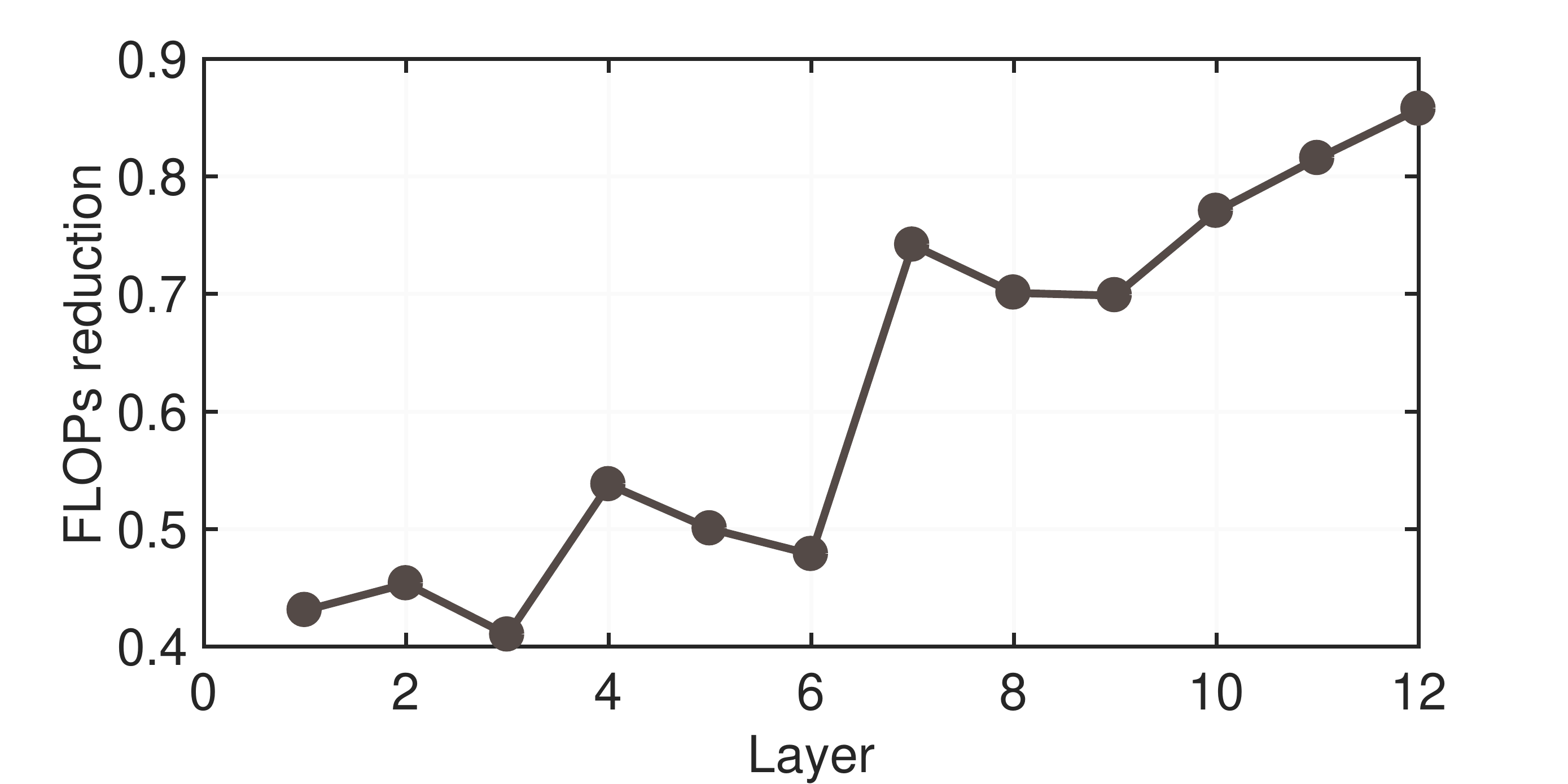}
\end{tabular}
\caption{\textbf{Left}: Ablation study on the effectiveness of multi-dimensional compression.
\textbf{Middle}: Layer-wise pruning ratio in our learned multi-dimensional pruning policy by GP search.
\textbf{Right}: Layer-wise FLOPs reduction.
All results are obtained with DeiT-B model on ImageNet.
\vspace{-0.1in}
}
\label{tab:ablation}
\vspace{-0.1in}
\end{table*}
\endgroup

\begingroup
\setlength{\tabcolsep}{1pt}
\begin{table}[t]
\centering
\footnotesize
\begin{tabular}{l | c c | c c | c c} \toprule
    \textbf{Model} & \multicolumn{2}{c|}{{DeiT-S}} & \multicolumn{2}{c|}{{DeiT-B}} & \multicolumn{2}{c}{{T2T-ViT-14}} \\
    & Base & Pruned & Base & Pruned & Base & Pruned \\\midrule
    \textbf{Top-1(\%)} & 79.8 & 79.9 / 79.3 & 81.8 & 82.3 / 81.5 & 81.5 & 81.7 \\ 
    \textbf{Throughput (img/s)} & 2773 & 4050 / 5523 & 1239 & 1792 / 2649 & 1940 & 2527 \\\bottomrule
\end{tabular}
\caption{Compare throughput of compressed models over baselines.
\vspace{-0.2in}}
\label{tab:throughput}
\vspace{-0.1in}
\end{table}
\endgroup


\begin{figure*}
\begin{floatrow}
\ffigbox[\FBwidth][]{%
\centering
\includegraphics[scale=0.16]{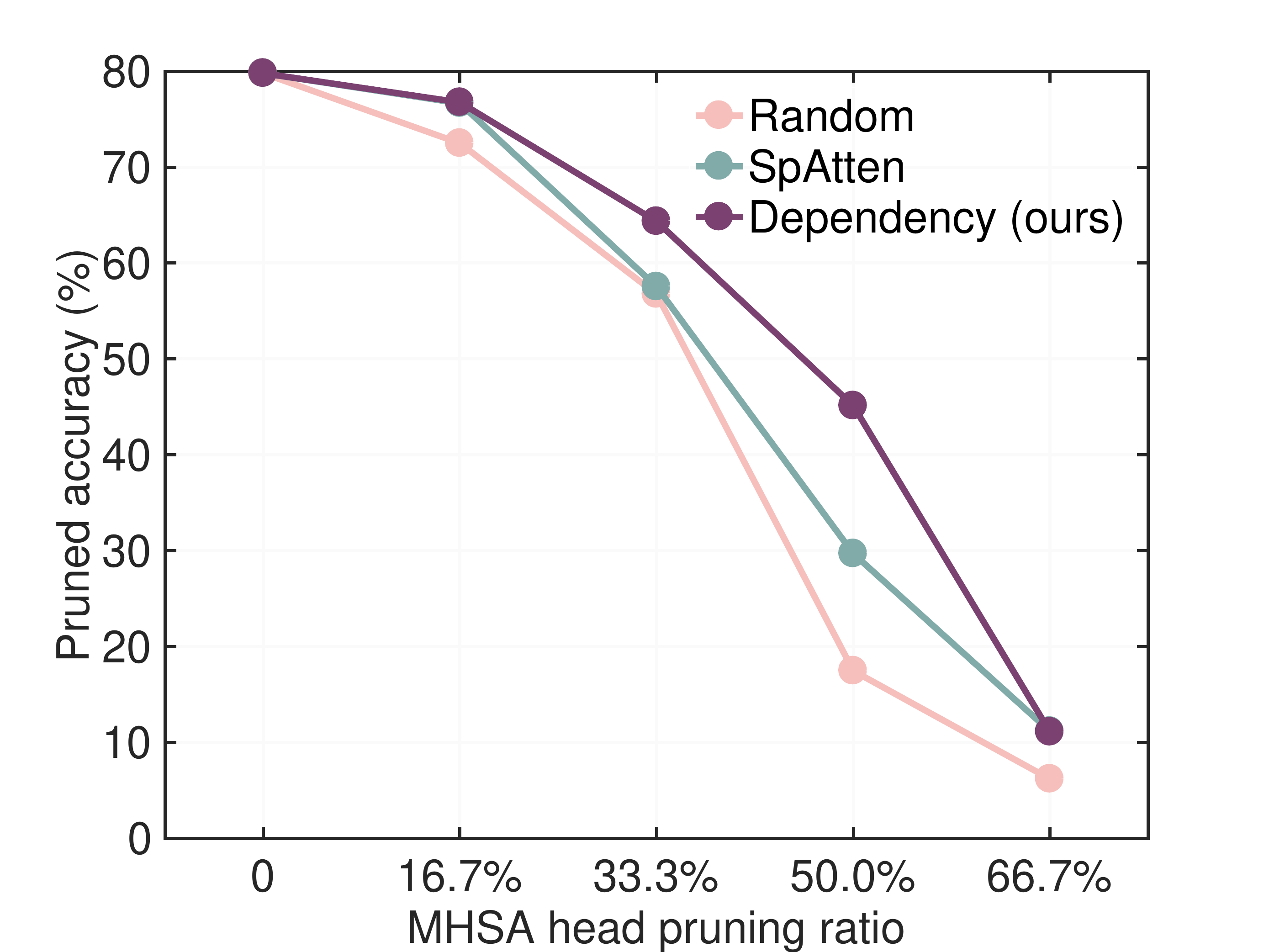}
\hspace{-0.2in}
\includegraphics[scale=0.16]{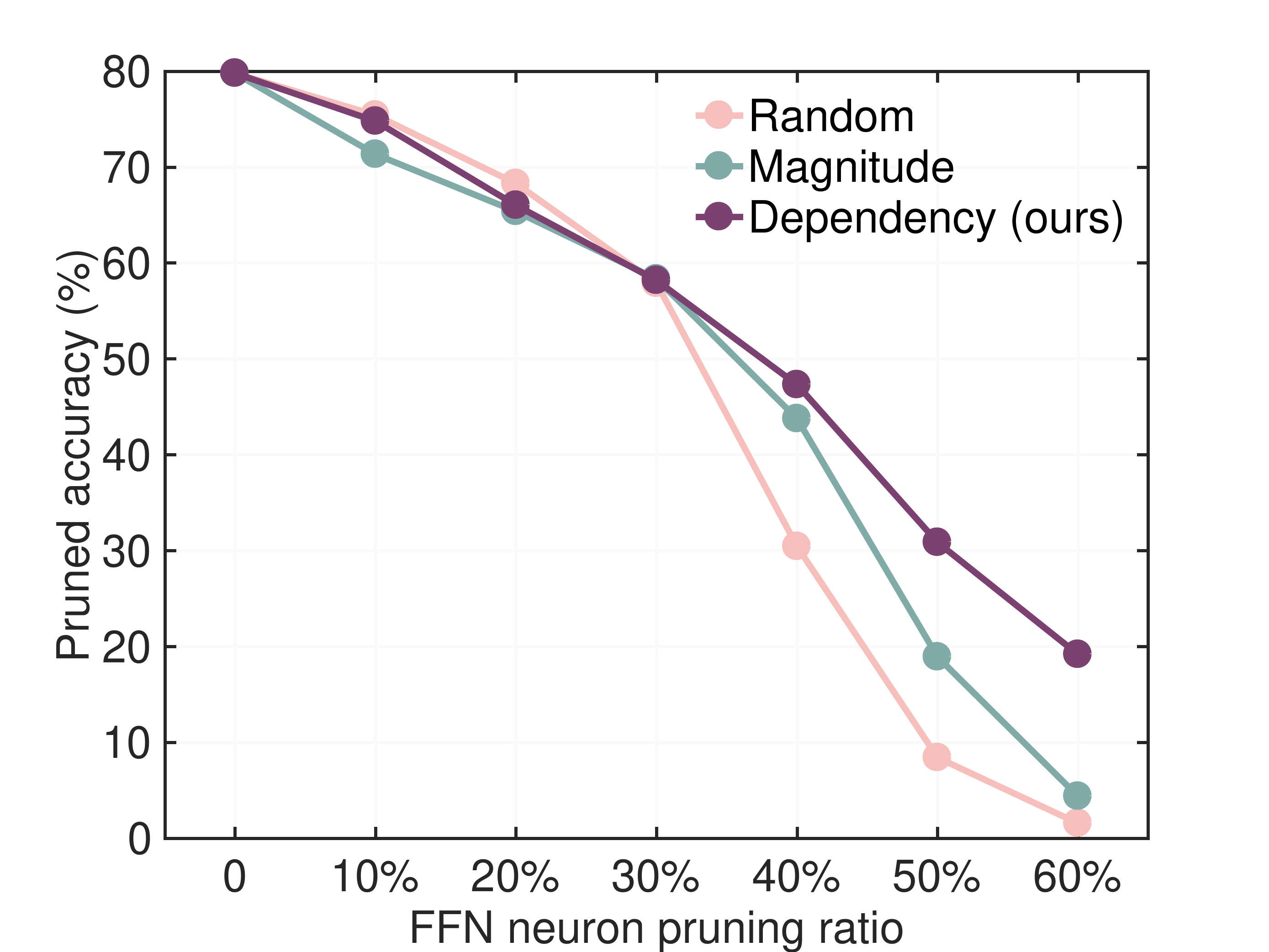}
\hspace{-0.2in}
\includegraphics[scale=0.16]{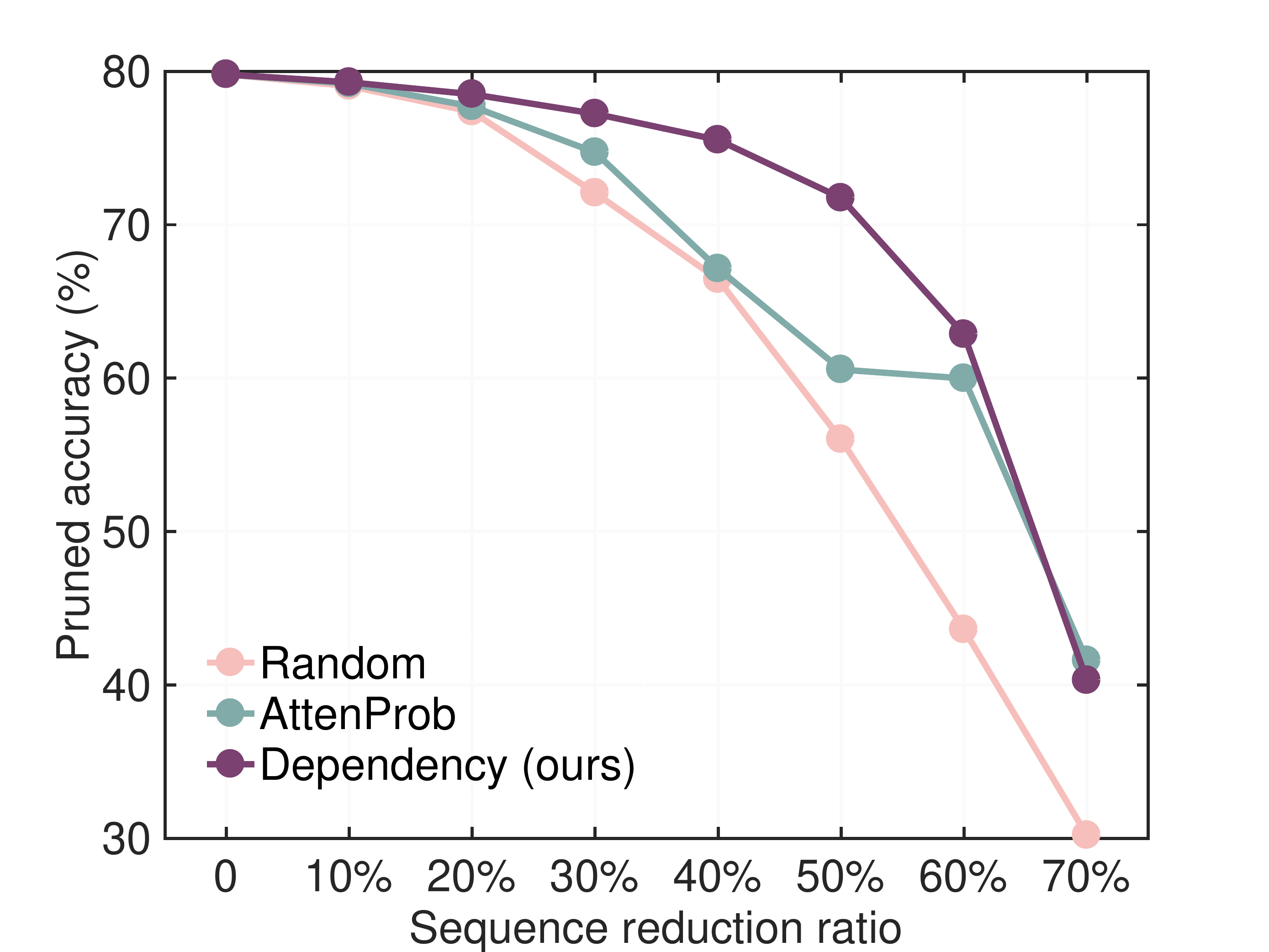}
}{%
  \caption{Compare the our dependency criterion versus other metrics for pruning each dimension individually. The plots show the pruned accuracy (without further finetuning) versus pruning ratios in each dimension. Results are obtained with DeiT-S on ImageNet.}%
}
\hspace{-0.2in}
\ffigbox[\FBwidth][]{%
\centering
\includegraphics[scale=0.16]{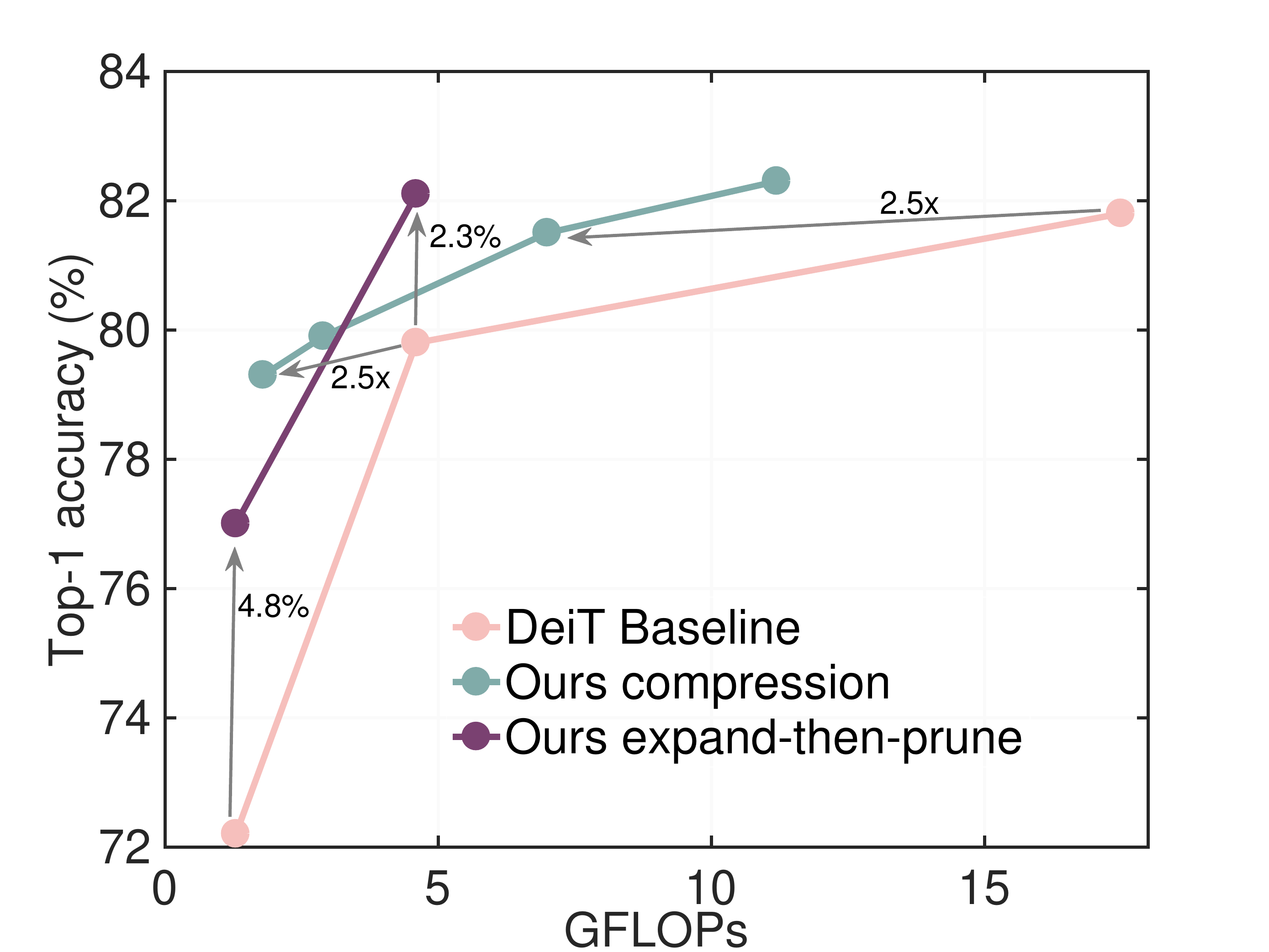}
}{%
  \caption{Comparison of the accuracy-FLOPs Pareto curve of our compressed models versus baseline.}%
}
\vspace{-0.1in}
\end{floatrow}
\end{figure*}

\begin{figure}[t]
\centering
\includegraphics[scale=0.2]{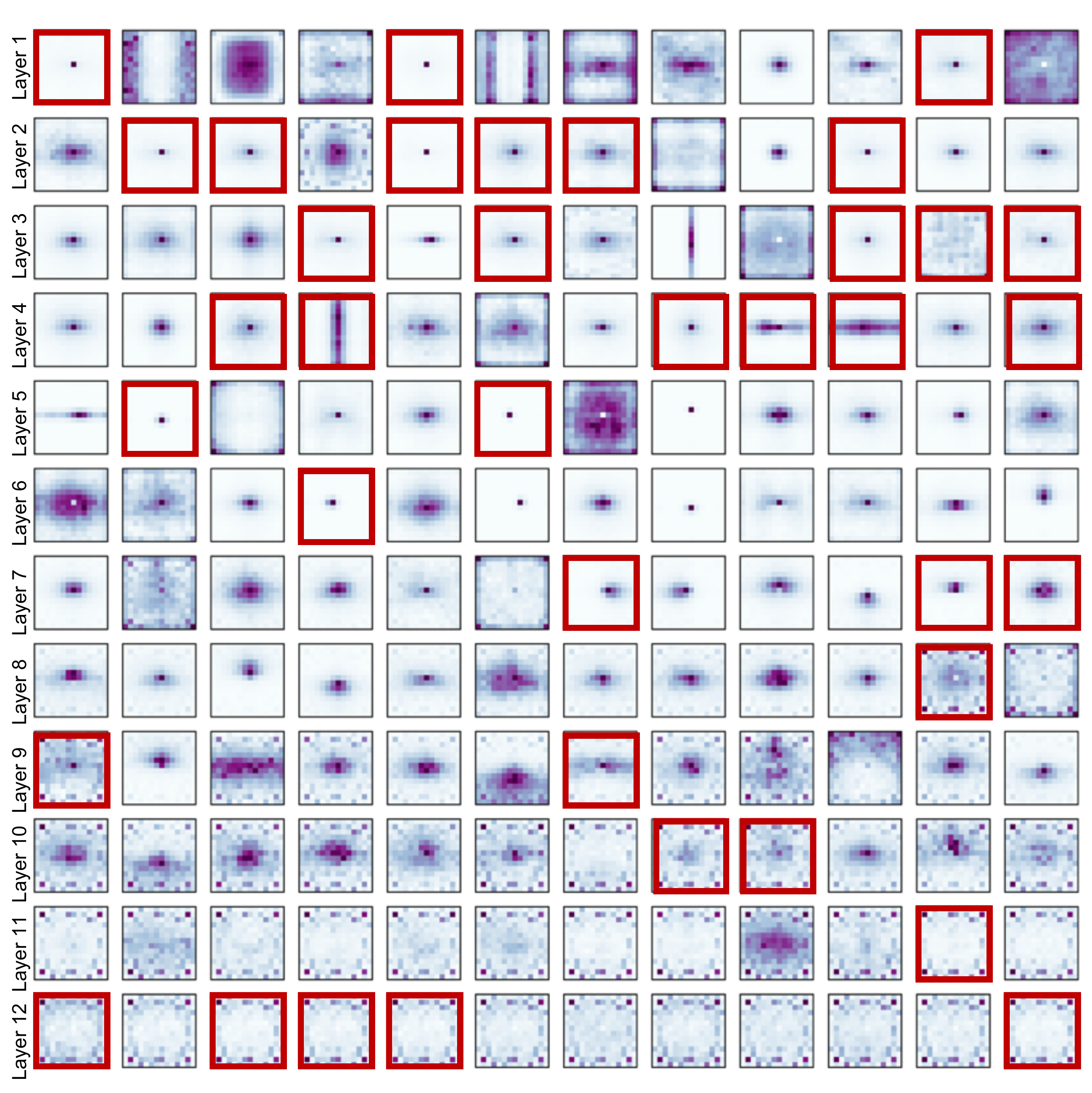}
\caption{Visualization of the attention-maps (averaged over 256 images) produced by all heads in the DeiT-B model. Red box means the head is pruned based on our dependency criterion. Number of heads removed follow our pruning policy in \Cref{tab:ablation}.
\vspace{-0.2in}
}
\label{fig:head_attention}
\vspace{-0.1in}
\end{figure}

\vspace{-0.1in}
\subsection{Ablation study.}
\label{sec:ablation}
\vspace{-0.1in}


\paragraph{Effect of multi-dimensional compression.}
We investigate the effectiveness of multi-dimensional compression in \Cref{tab:ablation}, where we compare with reducing the neuron, head or sequence individually for DeiT-B on ImageNet. 
Firstly, certain uni-dimensional compression method (e.g., head pruning alone) cannot yield significant FLOPs reduction, since the FLOPs of all the MHSA modules only account for 40\% of the total FLOPs. 
Secondly, although each dimension is prunable to some extent, 
excessive pruning of whichever dimension causes unacceptable accuracy loss, even for the fine-grained sequence reduction.
In contrast, 
our multi-dimensional compression achieves more FLOPs reduction with better accuracy. 
{Therefore, finding the optimal strategy to harness the FLOPs reduction from different dimensions is of vital importance.}
In \Cref{tab:ablation}, we also visualize our learned pruning policy by plotting the layer-wise pruning ratio and FLOPs reduction.
As shown, most head/neuron reductions come from shallower layers while most sequence reductions come from deeper layers.
Moreover, deeper layers may have more redundancy reflected by the increased layer-wise FLOPs reduction.



\paragraph{Actual inference speedup.}
We compare the throughput of our compressed models over baselines on a single Nvidia A100 GPU with a fixed batch size of 256. 
A shown in \Cref{tab:throughput}, our compressed models achieve $1.3\times \sim 2.2\times$ times throughput improvement without significant accuracy loss.

\paragraph{Different pruning criteria.}
In Figure 1, we compare our dependency based pruning criterion with previous metrics proposed for a specific pruning dimensions: SpAtten \cite{wang2021spatten} for head pruning, attention probability \cite{wang2021spatten} (AttenProb) for sequence reduction, and row-wise norm of the weight matrix (Magnitude) for neuron pruning.
All criteria (including random selection) perform well when the pruning rate is small ($<20\%$), suggesting that the redundancy indeed exists in each dimension. However, our dependency based pruning achieves relatively higher accuracy at larger pruning rates.
In \Cref{fig:head_attention}, by visualizing the attention-maps produced by all the heads in 
DeiT-B model, we observe that dependency based pruning indeed removes the redundant heads.

\begin{table}[t]
\centering
\footnotesize
\begin{tabular}{l l c c}\toprule
    \textbf{Model} & \textbf{Method} & \textbf{Top-1} & \textbf{FLOPs} \\\midrule
    \multirow{2}{*}{DeiT-S} & Random search & 76.4\% & 1.8G \\
    & GP search (ours) & \textbf{79.3\%} & 1.8G \\
    \bottomrule
\end{tabular}
\caption{Compare GP search with random search.
\vspace{-0.2in}
}
\label{tab:compare_random_search}
\vspace{-0.1in}
\end{table}

\paragraph{GP search versus Random search.}
Compared to random search \cite{li2020random} which determines the pruning policy by selecting the candidate with best validation accuracy 
from a random population, 
our GP search obtains better compressed model with 2.9\% higher accuracy at the same FLOPs (Table 5).

\vspace{-0.1in}
\section{Conclusion}
\label{sec:conclusion}
\vspace{-0.1in}

In this paper, we present a novel ViT model compression framework which prunes a pre-trained ViT model from attention head, neuron, and sequence dimensions jointly.
We firstly propose a statistical dependency based pruning criterion based on the Hilbert-Schmidt norm of the cross-covariance operator in order to identify the deleterious features in different dimensions.
Moreover, our framework learns the optimal pruning policy by casting multi-dimensional compression as an constrained non-linear optimization and using Gaussian process search with expected improvement to solve it.
Our results on ImageNet with various ViT models outperform previous state-of-the-art ViT pruning methods significantly under a comparable computational budget. 
One limitation is that our method does not explicitly incorporate model depth into the compression space. Instead, depth reduction is implicitly covered since pruning all the neurons or heads removes the layer except for the skip connection. 
However, this is never the case for our learnt pruning policy (Table 3), suggesting layer removal may be too aggressive for current ViT models.

{\footnotesize
\bibliographystyle{IEEEbib}
\bibliography{main}
}

\vspace{-0.1in}
\subsection*{Appendix}


\subsubsection*{A1.~~~Algorithm pseudo-code}
\vspace{-0.1in}

\begin{algorithm}[h]
\small
    \textbf{Input}: An $L$-layer ViT model with weights $\mathcal{W}=\{\mathbf{w}^1,...,\mathbf{w}^L\}$; pre-training iterations $T_{\text{pre}}$; target computational cost $\mathcal{T}$; population size $m$; GP search iterations $T_{\text{gp}}$; finetuning iterations $T_{\text{ft}}$; training set $\mathcal{D}_{\text{tr}}$; hold-out validation set $\mathcal{D}_{\text{val}}$  \;
    \textbf{Output}: Compressed ViT model satisfying the constraint $\mathcal{T}$ and its optimal weights $\mathcal{W}^*$ \;
    \tcc{Pre-training with Eq.(5)}
    Randomly initialize the model weights $\mathcal{W}$\;
    \For{each training iteration $t\in[T_\text{pre}]$}
    {
    Sample a mini-batch $(x,y)$ from $\mathcal{D}_{\text{tr}}$, sample a pruning policy $\omega$ from $U_{\mathcal{T}}$, and select weights $\mathcal{W}(\omega)$ by weight sharing as described in Sec.3.3 \;
    Compute training loss $\mathcal{L}(y|x;\mathcal{W}(\omega))$, backprop and update $\mathcal{W}$ \;
    }
    \tcc{GP search as described in Sec.3.3}
    Randomly sample $m$ different pruning policies $\{\omega_i\}_{i=1}^m$ satisfying $\mathcal{T}$, get $m$ compressed models (with weights $\mathcal{W}(\omega_i)$) by dependency based pruning as described in Sec.3.2, evaluate their actual accuracy $\mathcal{A}(\omega_i)$ on $\mathcal{D}_{\text{val}}$, and fit a GP model with $\Omega=\{\omega_i,\mathcal{A}(\omega_i)\}_{i=1}^m$ \;
    \For{each search iteration $t\in[T_\text{gp}]$}
    {
    Solve the non-linear programming Eq.(4) by SQP to get pruning policy $\omega_t^*$ \;
    Evaluate the actual accuracy of $\omega_t^*$ on $\mathcal{D}_{\text{val}}$\;
    Augment $\{\omega_t^*,\mathcal{A}(\omega_t^*)\}$ to $\Omega$ and refine the GP model \;
    }
    \tcc{Final pruning and finetuning}
    Compress the ViT model with the optimal pruning policy $\omega^*_{T_{\text{gp}}}$ using dependency based pruning \;
    Finetune the compressed model by $T_{\text{ft}}$ iterations \;
\caption{Multi-dimensional ViT compression.}
\label{algorithm: overview}
\end{algorithm}

\subsubsection*{A2.~~~More results of object detection}

For object detection, we apply our method to compress RetinaNet \cite{lin2017focal} with PVT-Small \cite{wang2021pyramid} backbone on COCO2017 dataset \cite{lin2014microsoft}.
Following \cite{wang2021pyramid}, models are trained on COCO train2017 (118k images) and evaluated on val2017 (5k images). 
We use the same training hyper-parameters as \cite{wang2021pyramid} to finetune the compressed model: AdamW optimizer with a batch size of 16, an initial learning rate of $1\times 10^{-4}$, standard 1$\times$ training schedule (12 epochs), learning rate is decayed by 10 at epoch 8 and 11.
The training image is resized to have a shorter side of 640 pixels, and during testing the shorter side of the input image is fixed to 640 pixels.


Considering that PVT-Small is an efficient ViT models and itself has progressive sequence shrinking strategy, we modify our sequence reduction strategy a little bit. The token selection layer no longer discard the unimportant tokens (those with bottom-ranking dependency scores).
Instead, they are directly skipped without participating any operations.
The important tokens selected based on our dependency scores participate into the MHSA or FFN operations, and the output will be concatenated with the skipped unimportant tokens.

The results are shown in \Cref{tab:object_detection}.
Our compressed model shows almost the same AP values compared to the baseline model, while achieving 30\% FLOPs reduction.

\begingroup
\setlength{\tabcolsep}{3pt}
\begin{table}[ht]
\centering
\footnotesize
\begin{tabular}{c | c | c c c c c c}\toprule
    \multirow{2}{*}{\textbf{Backbone}} & \textbf{FLOPs} & \multirow{2}{*}{\textbf{AP}} & \multirow{2}{*}{\textbf{AP$_{50}$}} & \multirow{2}{*}{\textbf{AP$_{50}$}} & \multirow{2}{*}{\textbf{AP$_{S}$}} & \multirow{2}{*}{\textbf{AP$_{M}$}} & \multirow{2}{*}{\textbf{AP$_{L}$}}  \\
    & \textbf{reduction} & & & & & & \\\midrule
    PVT-Small \cite{wang2021pyramid} & 0\% & 38.7 & 59.3 & 40.8 & 21.2 & 41.6 & 54.4 \\
    \rowcolor{LightCyan} Ours & 30\% & 38.6 & 58.8 & 40.6 & 21.0 & 41.5 & 54.4 \\
    \bottomrule
\end{tabular}
\vspace{-0.1in}
\caption{Results of RetinaNet with PVT-Small backbone on COCO2017 data for object detection task.
\vspace{-0.1in}
}
\label{tab:object_detection}
\end{table}
\endgroup

\subsubsection*{A3.~~~Training/finetuning hyper-parameters}

We follow the same training strategy as \cite{touvron2021training,Yuan_2021_ICCV} to pre-train and finetune DeiT and T2T-ViT models for our experiments on ImageNet.
The hyper-parameters are shown in \Cref{tab:hyper}.

\begin{table}[h]
\centering
\footnotesize
\begin{tabular}{l c c}\toprule
    Hyper-parameters & DeiT & T2T-ViT \\\midrule
    \;Epochs & 300 & 310 \\
    \;Warmup epochs & 5 & 5 \\
    \;Batch size & 1024 & 512 \\
    \;Learning rate & 1e-3 & 5e-4 \\
    \;LR decay & cosine & cosine \\
    \;Weight decay & 0.05 & 0.05 \\\midrule
    \;Label smoothing & 0.1 & 0.1 \\
    \;Dropout & 0 & 0 \\
    \;Stoch. depth & 0.1 & 0.1 \\\midrule
    \;Mixup prob. & 0.8 & 0.8 \\
    \;Cutmix prob. & 1.0 & 1.0 \\
    \;Erasing prob. & 0.25 & 0.25 \\
    \;Random Aug. & Yes & Yes \\\midrule
    \;Knowledge distillation & No & No \\
    \bottomrule
\end{tabular}
\vspace{-0.1in}
\caption{Training/finetuning hyper-parameters for our experiments on ImageNet.
\vspace{-0.1in}
}
\label{tab:hyper}
\end{table}

\subsubsection*{A4.~~~Other visualizations.}

\begin{figure}[h]
\centering
\includegraphics[scale=0.27]{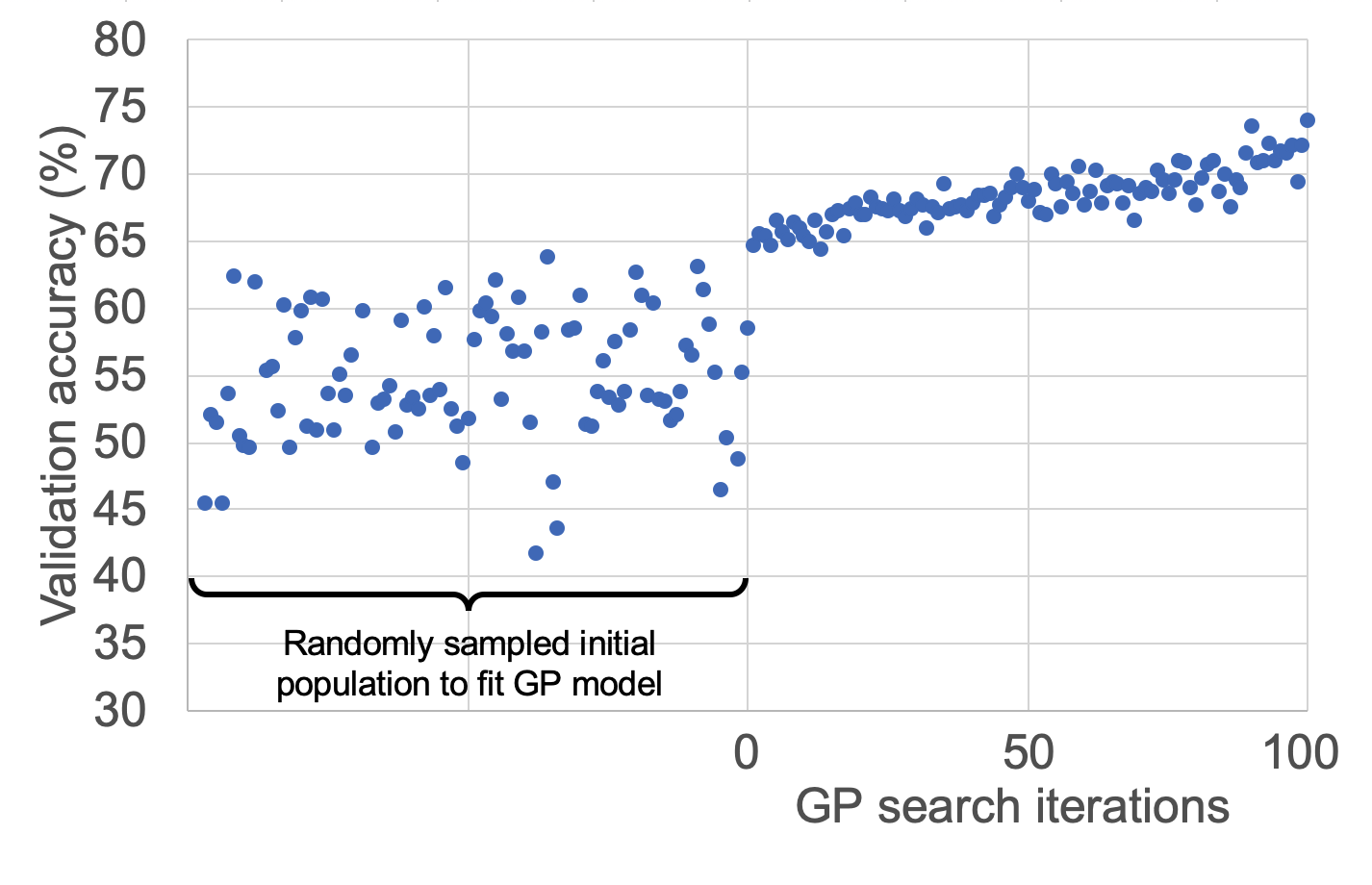}
\caption{GP search process with DeiT-B model on ImageNet by plotting the validation accuracy over search iterations.
{On the left: initial 100 randomly sampled population to fit GP model. On the right: accuracy of the pruning policy obtained by solving Eq.(4) at each search iteration.}}
\end{figure}

\begin{figure}[h]
\centering
\begin{subfigure}[b]{1.0\linewidth}
  \includegraphics[width=0.32\columnwidth]{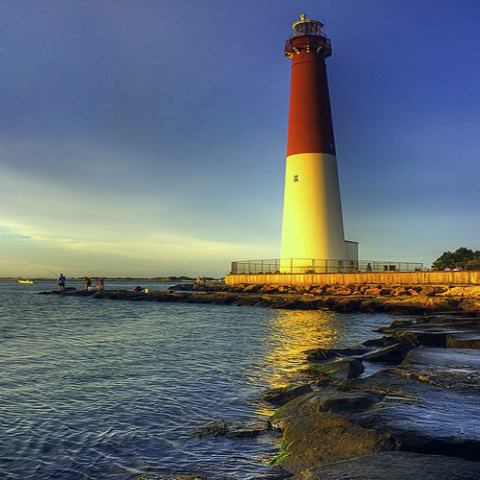}
  \includegraphics[width=0.32\columnwidth]{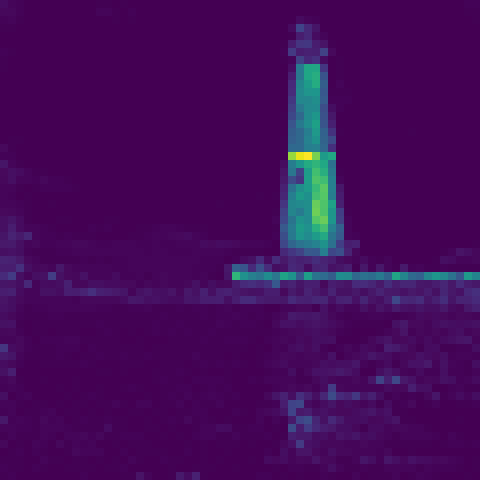}
  \includegraphics[width=0.32\columnwidth]{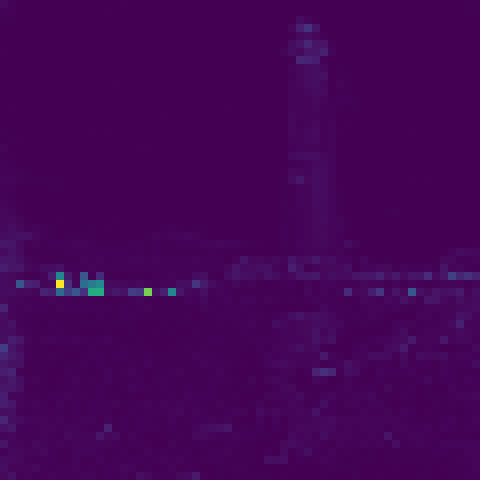}
\end{subfigure}
\begin{subfigure}[b]{1.0\linewidth}
  \includegraphics[width=0.32\columnwidth]{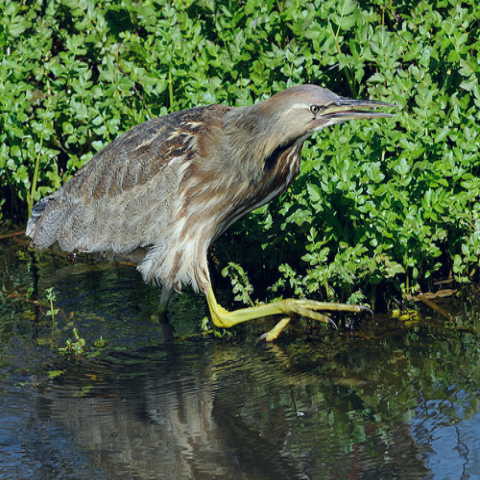}
  \includegraphics[width=0.32\columnwidth]{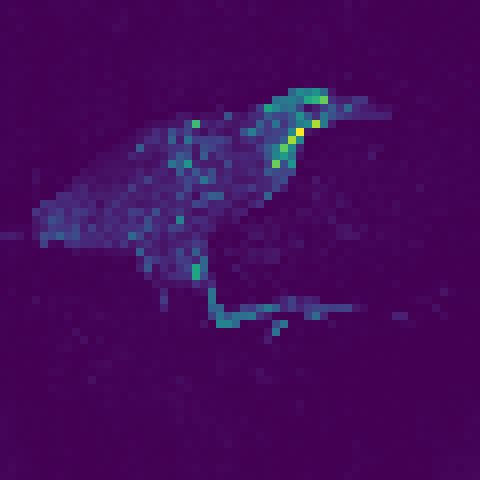}
  \includegraphics[width=0.32\columnwidth]{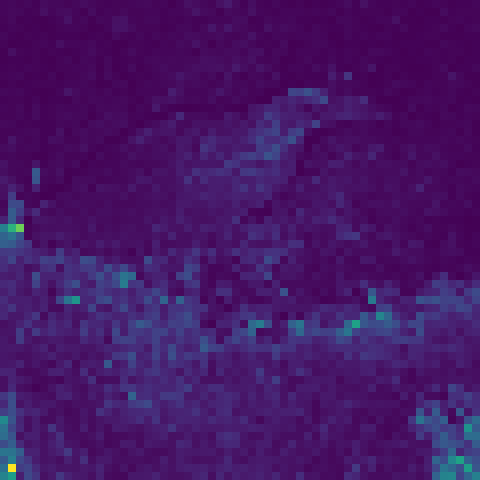}
\end{subfigure}
\begin{subfigure}[b]{1.0\linewidth}
  \includegraphics[width=0.32\columnwidth]{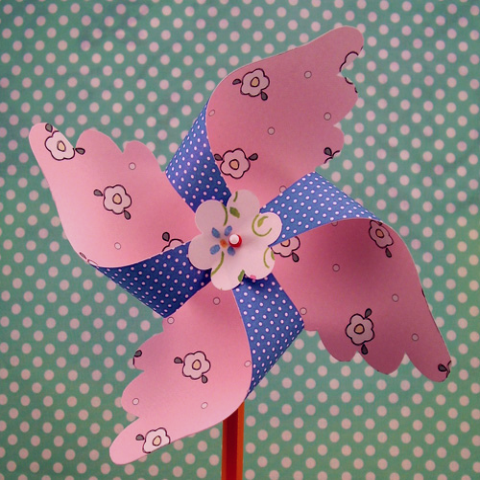}
  \includegraphics[width=0.32\columnwidth]{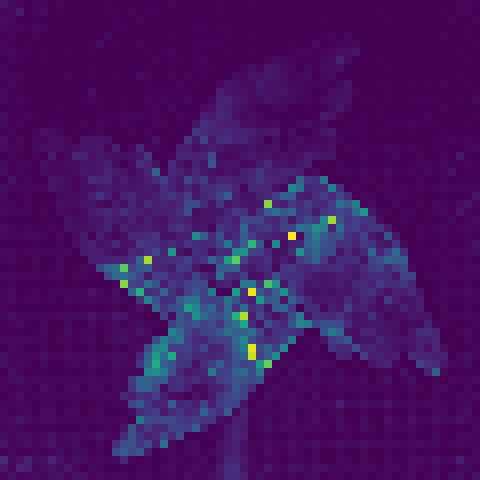}
  \includegraphics[width=0.32\columnwidth]{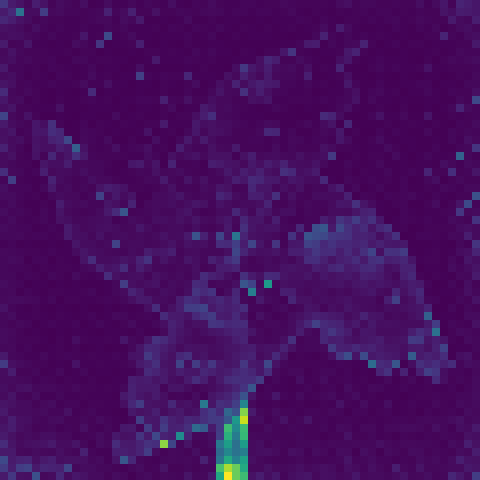}
\end{subfigure}
\centerline{{\small ~~~ \textbf{Image} ~~~~~~~~~~~~~~~~~ \textbf{High dep. head} ~~~~~~~~~~ \textbf{Low dep. head}}}
\caption{Examples of the attention-maps produced by top-ranking and bottom-ranking (in terms of our dependency criterion) attention head in the last block of DeiT-B model for several input images.
}
\end{figure}

\end{document}